\def\year{2021}\relax
\documentclass[letterpaper]{article} 
\usepackage{aaai21}  
\usepackage{times}  
\usepackage{helvet} 
\usepackage{courier}  
\usepackage[hyphens]{url}  
\usepackage{graphicx} 
\usepackage{tabularx}
\usepackage{makecell}
\usepackage{multirow}
\usepackage{algpseudocode}
\usepackage{booktabs}

\usepackage{bm}
\usepackage{natbib}  
\usepackage{caption} 
\usepackage{color}
\usepackage{xcolor}
\usepackage{amsmath}

\newcommand{\blue}[1]{{\color{blue} #1}}

\title{Generating Long Financial Report using Conditional Variational Autoencoders with Knowledge Distillation}
\author{
    Yunpeng Ren\textsuperscript{\rm 1,2}, 
    Ziao Wang\textsuperscript{\rm 1},
    Yiyuan Wang\textsuperscript{\rm 1},
    Xiaofeng Zhang\textsuperscript{\rm 1}
    \\
}
\affiliations{
    \textsuperscript{\rm 1}School of Computer Science, Harbin Institute of Technology Shenzhen, China\\
    \textsuperscript{\rm 2}Qianhai Financial Holdings Co., Ltd., Shenzhen, China\\

    Shenzhen University Town, Nanshan District, Shenzhen 518055\\
    zhangxiaofeng@hit.edu.cn

}
\begin{document}

\maketitle

\begin{abstract}
Automatically generating financial report from a piece of news is quite a challenging task. 
Apparently, the difficulty of this task lies in the lack of sufficient background knowledge to effectively generate long financial report. To address this issue, this paper proposes the conditional variational autoencoders (CVAE) based approach which distills external knowledge from a corpus of news-report data. Particularly, we choose Bi-GRU as the encoder and decoder component of CVAE, and learn the latent variable distribution from input news. A higher level latent variable distribution is learnt from a corpus set of news-report data, respectively extracted for each input news, to provide background knowledge to previously learnt latent variable distribution. Then, a teacher-student network is employed to distill knowledge 
to refine the output of the decoder component. To evaluate the model performance of the proposed approach, extensive experiments are preformed on a public dataset and two widely adopted evaluation criteria, i.e., BLEU and ROUGE, are chosen in the experiment. The promising experimental results demonstrate that the proposed approach is superior to the rest compared methods. 
\end{abstract}

\section{Introduction}\label{sec:intro}
Text generation \cite{2018QuaSE,duan2019pre-train} has long been investigated in the domain of natural language processing with flourishing results \cite{feng2018topic,serban2017multiresolution,mager2020gpt,hossain-etal-2020-simple,wang-etal-2019-topic-guided}. Generally, text generation task could have several sub problems, e.g., long text generation \cite{guo2017long,bahdanau2014neural,shen2019towards,dai2019transformer} and summary generation \cite{gao2020supert,genest2011framework,liu-lapata-2019-text,cao-etal-2018-retrieve}. Among these sub problems, long text generation from short text is quite challenging especially in a domain-specific task, such as financial report generation. 
Particularly, the difficulty to generate financial reports given a piece of short news lies in the lack of sufficient information, especially the financial background knowledge, and thus this challenging task is seldom addressed by existing work. 

\textbf{Prior works.} In the literature, there exist a good number of recurrent neural network (RNN) based text generation approaches \cite{feng2018topic,sorodoc2020probing,xu2020self} although not for long text generation \cite{guo2017long}. Generally, these approaches can be classified into two categories, i.e., \textit{deep generative model based approaches} and \textit{generative  adversarial network (GAN) based approaches}. As for the generative model based approaches, variational autoencoder (VAE) approach is widely adapted by many researchers for this task \cite{kingma2013auto,mccarthy2020addressing,2019T}. In these VAE based approaches, LSTM or GRU-like component is usually chosen as the encoder or decoder component. Also, the conditional variational autoencoder \cite{2018Generating} with a hybrid decoder adding the deconvolutional neural networks was proposed 
to learn topics to generate Chinese poems. Particularly, \cite{shen2019towards} designed a hierarchy of stochastic layers between the encoder and decoder component to learn a VAE model for generating long coherent text. For GAN based approaches, a good number of research attempts \cite{2019Enhancing,guo2017long,zhang-etal-2020-improving-adversarial} have been made towards generating high-quality long text. To incorporate background knowledge, \cite{2019Enhancing} extracts knowledge from external base to integrate with 
the generator through a dynamic memory mechanism, and the model is adversarial trained with a multi-classifier as the discriminator. 


However, there exist two challenges 
which invalidate most existing approaches. First, the length of the input eventual news is rather short, e.g., “The coronavirus recession struck swiftly and violently. Now, with the US economy still in the grip of the outbreak five months later, the recovery looks fitful and uneven and painfully slow”. But the length of the generated reports is usually much greater than that of the input news, and apparently it is quite challenging to generate reasonable reports. 
Second, the generation of financial reports by human specialists often involves their 
intellectual efforts, such as inferring and reasoning abilities. Till now, these challenges remain unresolved 
and need more and more research efforts. 

\begin{figure}[t]
\centering
\includegraphics[width=1\columnwidth]{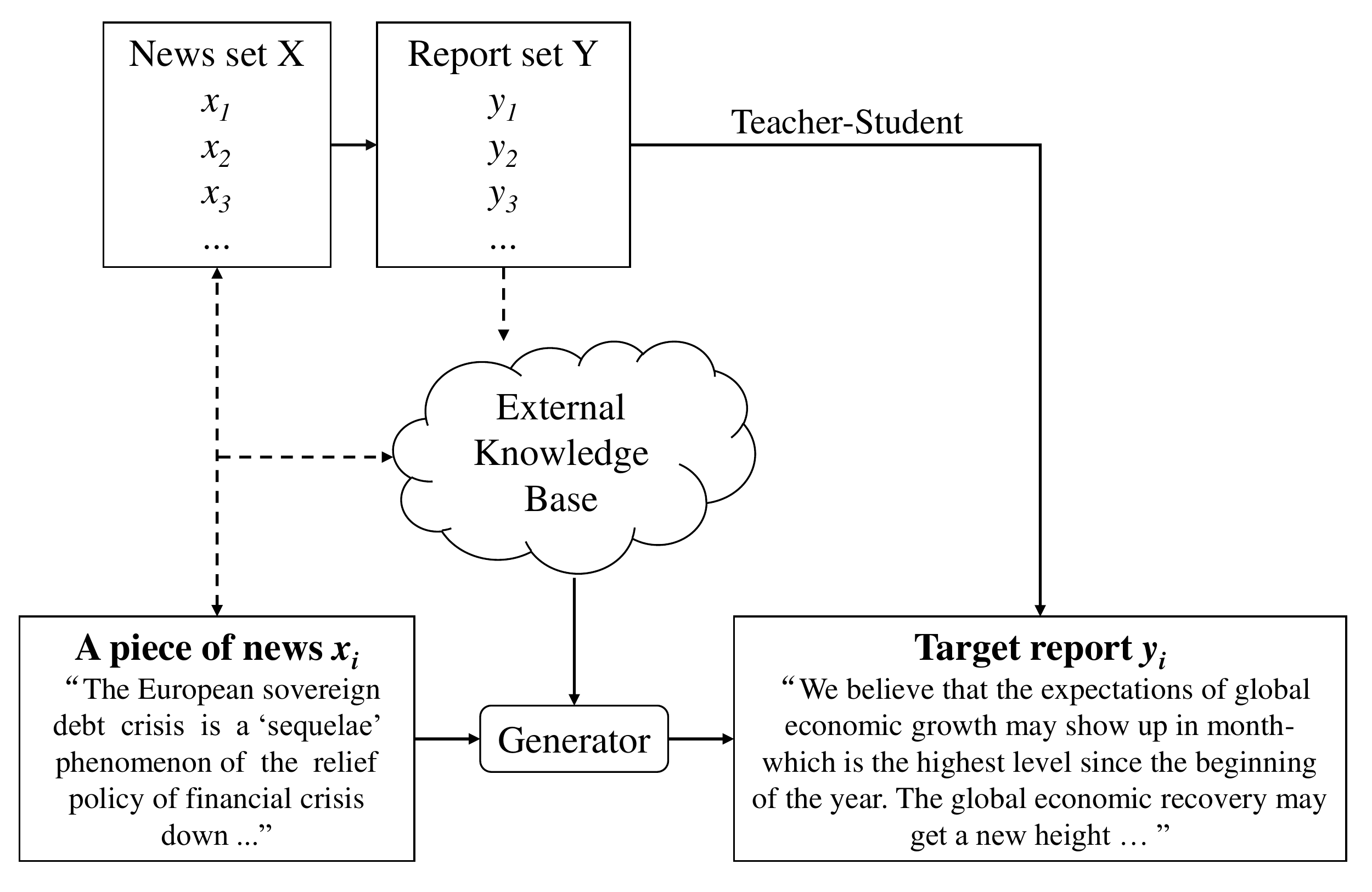} 
\caption{An illustrating example of the proposed approach. }
\label{fig:illustration}
\end{figure}

To address 
aforementioned issues, we propose this novel conditional variational autoencoders based approach with knowledge distillation, called CVAE-KD. An illustrating example of the proposed approach is plotted in Figure \ref{fig:illustration}. In the proposed approach, 
a carefully designed teacher-student network structure is integrated with the CVAE model to simultaneously resolve the information loss as well as the knowledge reasoning issues. To encode input news data, a Bi-GRU component is adopted and the latent variable distribution is learnt on each batch of input news data. Obviously, the learnt latent variable $z_1$ does not contain sufficient information to decode a high quality report. To provide sufficient information as well as to simulate human's inference knowledge, a higher level latent variable distribution $z_2$ is learnt from a corpus of historical news-report data. To embed these data, a pre-trained component ELMO \cite{peters-etal-2018-deep} is adopted and then $z_2$ is estimated using the feature embeddings of aggregated similar news data. During model learning, $z_1$ is forced to approximate $z_2$ by a designed KL-divergence term. For decoder component, a GRU component is adopted. To further refine the output reports, a set of financial reports of similar news are embedded through the same ELMO component, and are then treated as a teacher. The output of decoder component is treated as a student component. With this knowledge distillation step, the corresponding financial report for a piece of input news is already generated.

The main contributions of this paper are summarized as follows. 
\begin{itemize}
    \item We propose the conditional variational autoencoders (CVAE) based approach with knowledge distillation to generate financial reports. To the best of our knowledge, this is among the first attempts to resolve this challenging task. 
    
    \item We employ a pre-trained model as a teacher network and the student component is to learn background knowledge from external knowledge base. The corresponding KL-divergence loss is designed for this purpose. 
    
    \item Extensive experiments are evaluated on one real-world dataset. The promising experimental results have demonstrated the superiority of the proposed approach over the compared baseline and the state-of-the-art approaches. 
\end{itemize}


\section{The Proposed Approach}
\begin{figure*}[t]
\centering
\includegraphics[height=8cm, width=0.9\textwidth]{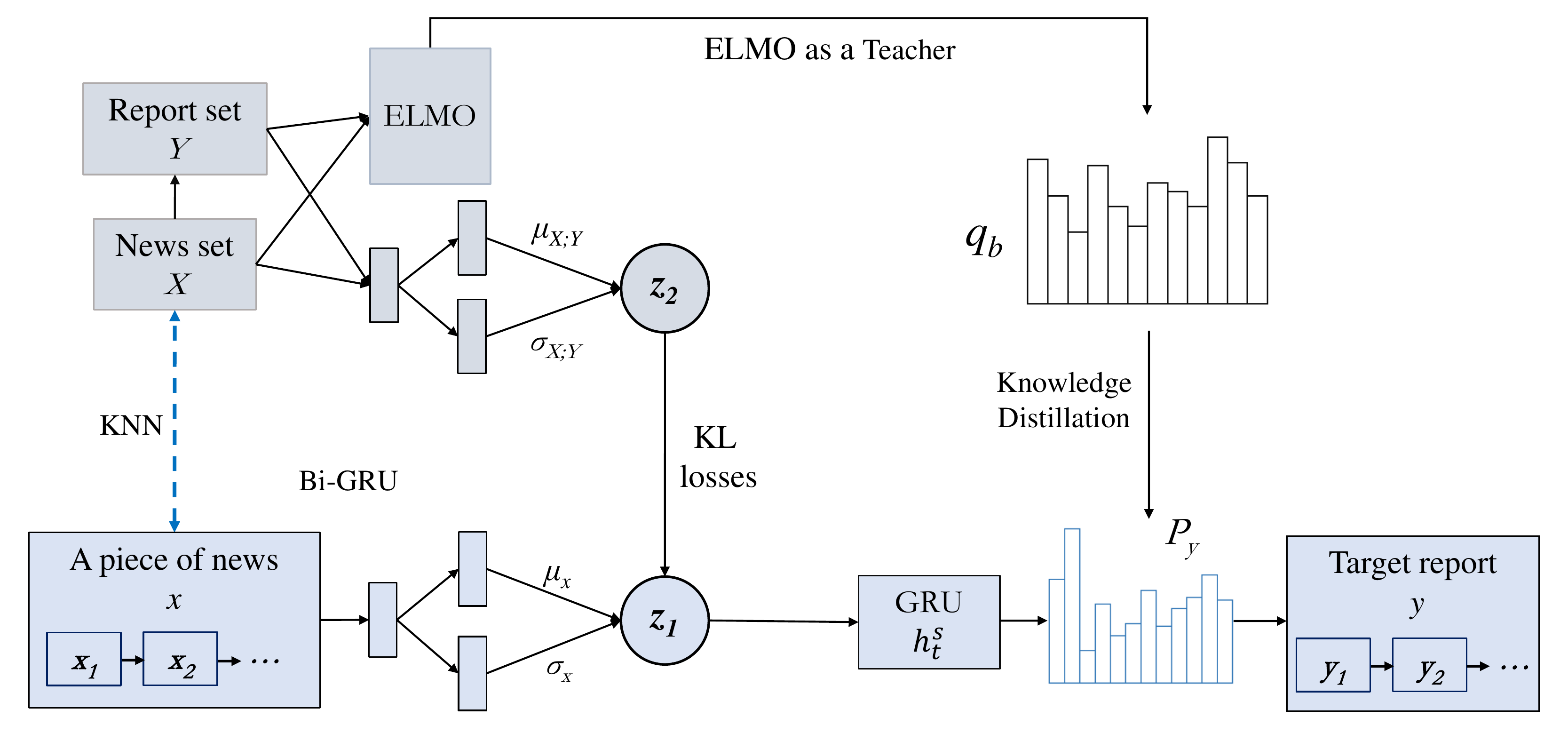} 
\caption{Details of the proposed CVAE-KD.}\label{fig:framework}
\end{figure*}
\subsection{Preliminaries and Problem Setup}
Let $X$ denote a corpus of news data with each $x = (x_1,...,x_M)$ containing $M$ tokens, $Y$ denote the set of generated reports and each $y = (y_1,...,y_N)$ containing $N$ tokens, where $N$ is significantly greater than $M$. The financial report generation problem could be formulated as 
\begin{equation}\label{eq:problem}\nonumber
    p(y_1,y_2,...,y_N) = \prod_{k=1}^N p(y_k|y_{1:k-1};x).
\end{equation}
That is, given a piece of news $x$, together with the generated $k-1$ tokens $y_{1:k-1}$, we predict the probability of the next token to be generated. 

\subsection{The proposed CVAE-KD}
The framework of the proposed CVAE-KD model is plotted in Figure \ref{fig:framework}. In this approach, the input news $X$ is first embedded using lookup table. Then, Bi-GRU is adopted to encode each input $x$, and the latent variable distribution is learnt. For the background knowledge learning, a set of similar news are extracted as well as the corresponding reports to estimate a higher level latent variable distribution. At last, a GRU component is employed to decode the output report. The proposed CVAE-KD has three components, i.e., encoder component, background knowledge extraction component, decoder component, and we detail them in the following subsections. 

\subsection{Encoder Component}
We adopt a Bi-GRU component for the encoder component. The input of this component is a piece of news $x$. First, $x$ is embedded using lookup table and then is fed into the Bi-GRU component. As usual, we only choose the hidden state $h_t^e$ of Bi-GRU as the output of this component, the general calculation of Bi-GRU are given as 
\begin{eqnarray}
\nonumber c_t &=& \sigma(W_c x + U_c h_{t-1}^{ex})\\
\nonumber r_t &=& \sigma(W_r x + U_r h_{t-1}^{ex})\\
\nonumber h_t^{x^`} &=& tanh(W x + U(r_t  h_{t-1}^{ex})\\
\nonumber h_t^{xe} &=& (1-c_t) h_{t-1}^{ex} + c_t h_t^{x^`}\\
\nonumber h^{e} &=& [\overrightarrow{h_t^{xe}}, \overleftarrow{h_t^{xe}}],
\end{eqnarray}
where $c_t$ is the update gate, $r_t$ is the reset gate,$h_{t-1}^{ex}$ is the previous activation, $h_t^{x^`}$ is the candidate activation, $h_t^{xe}$ is the current activation and $h^e$ is the concatenation of activation from both direction. 

After embedding news data into feature vectors. We assume the latent variables $z$ of CVAE follow a gaussian distribution. Then, two MLPs are respectively applied to learn parameters of the latent variable distribution, i.e. mean $\mu$ and standard deviation $\sigma$, and they are calculated as  
\begin{eqnarray}
\nonumber \mu_x &=& f_{\mu_x}(h_t^e) \\
\nonumber \sigma_x &=& f_{\sigma_x}(h_t^e). 
\end{eqnarray}
With the learnt latent variable distribution, we can then sample latent variable $z_1$ from it, written as
\begin{equation} \nonumber
    z_1^i = \mu_x^i + \sigma_x^i \epsilon, \epsilon \sim N(0,I)/N(\mu, \sigma).
\end{equation}

\subsection{Background Knowledge Extraction and Distillation}
\subsubsection{Learning latent variable from background Knowledge}
In this component, we first build external knowledge base for each input $x$. To this end, we apply the standard KNN algorithm to group a subset $X_s$ of news which are most similar to $x$. Then, we extract the set $Y_s$ of corresponding financial reports. 

Similar to the previous steps, we also first lookup $X_s$, $Y_s$ and then embed them using Bi-GRU component. Then, another gaussian distribution is assumed for this background knowledge base with the parameters are estimated as 
\begin{equation} \nonumber
    \mu_{X_{em};Y_{em}} = f_{\mu_x}([X_{em};Y_{em}])
\end{equation}
\begin{equation} \nonumber
    \sigma_{X;Y} = f_{\sigma_x}([X_{em};Y_{em}])
\end{equation}
\begin{equation} \nonumber
    z_2^i = \mu_{X;Y}^i + \sigma_{X;Y}^i \epsilon_{X;Y}, \epsilon_{X;Y} \sim N(0,I)/N(\mu, \sigma)
\end{equation}
where $X_k$ and $Y_k$ are feature representations globally learnt, $X_{em}$ and $Y_{em}$ feature representations of each input pair of data (news-report), $z_2$ is the latent variable sampled from the learnt gaussian distribution. 

\subsubsection{Knowledge distillation to supervise report generation}
To distill knowledge from previous extracted external knowledge base, the extracted subsets $X_s$ and $Y_s$ are first embedded using a pre-trained model ELMO. Its output are denoted as $ELMO_k^{task}$. Without loss of generality, we concatenate these ELMO embeddings with the original feature representations, and thus the output of this module can be written as 
\begin{eqnarray}
\nonumber    X_{em} = [X_k; ELMO_k^{task}] \\
\nonumber    Y_{em} = [Y_k; ELMO_k^{task}]. 
\end{eqnarray}
To align the output of this component with the vocabulary, a two-layers MLP is adopted. Then, $Y_{em}$ is considered as the teacher to supervise the generation of financial report $y$ given $x$. 

Accordingly, the student is the output of decoder component during model training process. Given the output of decoder, i.e., $y = (y_1, y_2,...,y_N)$, the employed ELMO is to predict probability of next token to be generated given a sequence of generated tokens in $y$, written as 
\begin{equation} \nonumber
    p(y_1, y_2, ..., y_N) = \prod_{k=1}^N p(y_k|y_1,y_2,...,y_{k-1})
\end{equation}
\begin{equation} \nonumber
    p(y_1, y_2, ..., y_N) = \prod_{k=1}^N p(y_k|y_{k+1},y_{k+2},...,y_{N}).
\end{equation}
The knowledge distillation is optimized according to following objective function, given as 
\begin{equation} \nonumber \label{eq:kdloss}
    L_{kd}(\theta) = -\sum_{w \in V}[P_{\phi}(y_t=\omega|x,y) 
    log P_{\theta}(y_t=\omega|x,y_{1:t-1})]
\end{equation}
where $P_{\phi}(y_t)$ is the soft target of ELMO, 
$\phi$ denotes the parameters of ELMO, and $V$ denotes the output vocabulary.

\subsection{Decoder Component}
To generate the report $y$, a GRU is chosen as the decoder component. The output of GRU is used as the output of this component, given as 
\begin{equation} \nonumber
    h_t^s = GRU(z_1).
\end{equation}
This output is then input to a MLP to generate the probability of word in adopted vocabulary, given as
\begin{equation} \nonumber
    p(y_k) = softmax(tanh(W_m h_t^s + b)).
\end{equation}
The word with the maximum probability value is chosen as the final output at each iteration. 

\subsection{Model Loss}
The overall loss function of the proposed CVAE-KD contains two terms, i.e., CVAE loss and knowledge distillation loss. Each loss item is respectively illustrated as follows. 

The CVAE loss contains two part, the first of which is to calculate the reconstruction loss of the model and the second is to force the latent variable $z_1$, learnt from input news $x$, to approximate the latent variable $z_2$, learnt from background knowledge base, written as
\begin{eqnarray} 
    \nonumber L_{CVAE}(x,y;\theta,\phi)  = -D_{KL}[q_{\varphi}(z_2|X,Y) || p_{\theta}(z_1|x)]\\
    \nonumber + E_{q_{\phi}(z_1|x, z_2)}[log p_{\theta}(y|z_1,x)]
\end{eqnarray}

where $p_{\theta}(z_1|x)$ , $q_{\phi}(z_1|x, z_2)$ and  $q_{\varphi}(z_2|X,Y)$ are respectively calculated as 
\begin{equation} \nonumber
    q_{\phi}(z_1|x, z_2) = N(z_1;\mu_{x},\mu_{z_2},\sigma_{x},\sigma_{z_2})
\end{equation}
\begin{equation} \nonumber
    p_{\theta}(z_1|x) = N(z_1;\mu_{x},\sigma_{x}). 
\end{equation}
\begin{equation} \nonumber
    q_{\varphi}(z_2|X,Y) = N(z_2;\mu_{X,Y},\sigma_{X,Y}).
\end{equation}

As the knowledge distillation loss is given in Eq. \ref{eq:kdloss}, the overall model loss of the proposed CVAE-KD can be written as 
\begin{equation} \nonumber
    L_{total} = \alpha L_{CVAE}  + (1 - \alpha)L_{kd}
\end{equation}
where $\alpha$ is a learnable parameter. The model is then optimized using Adam algorithm \cite{adam}. 

\begin{table*}[t]
\centering
\begin{tabular}{p{5cm} p{1.5cm} p{1.5cm} p{1.5cm} p{1.5cm}} 
\toprule[2pt] 
\centering
{Methods} & \centering{BLEU-1} & \centering{BLEU-2} & \centering{BLEU-3} & \,{BLEU-4}
\\
\hline
\end{tabular}
\begin{tabular}{p{5cm} p{1.5cm} p{1.5cm} p{1.5cm} p{1.5cm} }
\centering{Seq2seq} & \centering{32.69} & \centering{7.65} & \centering{4.85}& \quad{2.75} 
\\ 
\centering{Seq2seq+Attn} & \centering{33.64}  &\centering {13.85} &\centering {9.89} & \quad{6.92}\\ 
\centering{Pointer-Generator} & \centering{36.45}  & \centering{9.51} & \centering {5.75} & \quad{2.45}\\ 
\centering{CVAE} & \centering{33.5}  & \centering{14.07} & \centering{10.04}& \quad{6.97}
\\
\hline 
\end{tabular}
\begin{tabular}{p{5cm} p{1.5cm} p{1.5cm} p{1.5cm} p{1.5cm} }

\centering{\textbf{CVAE-KD}} & \centering{\textbf{46.67}}  & \centering{\textbf{20.32}} & \centering{\textbf{12.81}}& \quad{\textbf{8.00}}\\
\bottomrule[2pt]
\end{tabular}
\caption{Evaluation results of all compared methods with respect to BLEU criteria.}
\label{table1}
\end{table*}
\begin{table*}[t]
\centering
\begin{tabular}{p{5cm} p{2cm} p{2cm} p{2cm} }
\toprule[2pt] 
\centering
{Methods} & \centering{ROUGE-1} & \centering{ROUGE-2} & \,{ROUGE-L} \\
\hline
\end{tabular}
\begin{tabular}{p{5cm} p{2cm} p{2cm} p{2cm}  }
\centering{Seq2seq} & \centering{8.46} &  \centering{1.30}& \,\,\quad{3.59} 
\\ 
\centering{Seq2seq+Attn} & \centering{15.66} &\centering {3.02} & \,\,\quad{3.89}\\ 
\centering{Pointer-Generator} & \centering{12.08}  & \centering {1.40} & \,\,\quad{3.44}\\ 
\centering{CVAE} & \centering{16.69}  & \centering\textbf{{3.32}} & \,\,\quad{4.65}
\\
\hline
\end{tabular}
\begin{tabular}{p{5cm} p{2cm} p{2cm} p{2cm}  }
\centering{\textbf{CVAE-KD}} & \centering{\textbf{18.27}} & \centering{2.64}& \,\,\quad{\textbf{6.95}}\\
\bottomrule[2pt]
\end{tabular}
\caption{Evaluation results of all compared methods with respect to ROUGE criteria.}
\label{table2}
\end{table*}

\section{Experiments}
In this section, we first illustrate how the dataset is prepared as well as the evaluation criteria. Then, several baseline models as well as the SOTA approaches are introduced. At last, we perform extensive experiments on one real-world dataset to answer following research questions:
\begin{itemize}
    \item RQ1: Does the proposed approach outperform the state-of-the-art approaches for long text generation, i.e., financial reports, given a piece of short news?
    \item RQ2: What is the model performance when generating reports with different size? 
    \item RQ3: How is the quality of generated financial reports?
    \item RQ4: Whether the proposed knowledge distillation component could affect model performance (ablation study)?
\end{itemize}


\subsection{Dataset and Evaluation Criteria}
\subsubsection{Dataset Preparation} To evaluate model performance, we crawled financial news as well as the corresponding reports from these famous Chinese financial Websites (Sina Finance\footnote{http://stock.finance.sina.com.cn/stock/}, Tonghuashun Finance\footnote{http://m.10jqka.com.cn/ybnews/} and Eastmoney\footnote{http://data.eastmoney.com/report/}). The raw dataset 
contains 10,706 pairs of news-report data and each piece of news is associated with a financial report. 
\subsubsection{Data Pre-processing} An open source tool (“jieba”) \footnote{https://github.com/fxsjy/jieba} is first adopted to segment the Chinese news. After word segmentation, the average length of the financial news and the financial reports are 28 and 331 words, respectively. Then, we screen out the word 
if its term frequency (TF) is higher than 5, 
and the new vocabulary set contains 17,210 words. At last, we replaced the numeric symbols with number token (NUM). Furthermore, the vocabulary is marked with four other tokens, i.e., padding token (PAD), unknown token (UNK), start position token (START) and end position token (END).




\subsubsection{Evaluation criteria} The Bilingual Evaluation Understudy (BLEU) and ROUGE are chosen as the evaluation criteria. As most Chinese phases consist of less than five words, we chose the BLEU-1, BLEU-2, BLEU-3 and BLEU-4 scores as the detailed evaluation measurements. Similarly, we chose ROUGE-1, ROUGE-2 and ROUGE-L scores as the ROUGE-type evaluate criteria. 

\subsection{Baseline Models}
To evaluate the model performance, several baseline and the state-of-the-art approaches, i.e., Seq2Seq, Seq2Seq+Attn, Pointer-Generator network and CVAE model  
are chosen in the experiments for performance comparison. Details of each compared approach are illustrated as follows. 

\begin{itemize}
    \item Seq2Seq \cite{2014Sequence} is considered as a baseline model for text generation task which already achieves a superior model performance in various text-to-text generation problem.
    
    \item Seq2Seq+Attn \cite{bahdanau2014neural} extends the original Seq2Seq by allowing the soft-search for relevant words, from the input source sentences, to predict the next word to be generated. 
    
    \item Pointer-Generator network \cite{see2017get} is considered as the state-of-the-art model proposed for sampling words from the input source sentences via the pointing process, and then it integrates the coverage mechanism to 
    penalize the generation of repetitive words.  
    
    \item  CVAE \cite{zhao2017learning} is most related to our proposed approach. This model captures the discourse-level diversity in the encoder and uses greedy decoders to generate diverse responses. And the latent variables are used for learning the distribution over potential conversational intentions.
\end{itemize}

\subsection{Experimental Settings}
\subsubsection{Network Parameters}
The parameters of designed neural network model are designed as follows. For the employed GRU component, the number of hidden units in the GRU component is set to 256. The dimension of the hidden variables is set to 16. The dropout rate for decoder is set to 0.5 to avoid over-fitting problem. The learning rate is set to 0.001, the batch size is set to 16 and the weight for knowledge distilling loss is set to 1. 

\subsubsection{Experimental Settings}
In the experiments, the maximum length of encoder is set to 30. To generate reports of different length, the length of decoder is respectively set to 100, 150 and 200 to evaluate the model performance of the proposed CVAE-KD. Then, we evaluate all models and report the experimental results in following subsections. 

\subsection{Results on generating financial reports (RQ1)}
In this experiment, we 
evaluate all approaches and the corresponding BLEU and ROUGE scores are reported in Table \ref{table1} and Table \ref{table2}, respectively. 

From Table \ref{table1}, it is obvious that the proposed CVAE-KD is the best model. The corresponding BLEU-1, BLEU-2, BLEU-3 and BLEU-4 scores of CVAE-KD are 28.0$\%$, 44.4$\%$, 27.6$\%$, 14.8$\%$ higher than the second best scores, respectively. Except for BLEU-1 indicator, the CVAE is the second best model which partially verifies that the effectiveness of the proposed CVAE-KD over the CVAE model. It is also noticed that the Pointer-Generator (PG) network is the worst model. The possible reason is that this PG network is for summarization task from long text to short text, and it might not suit for our problem. We also observed that when more terms are generated together, the model performance, e.g., BLEU-4, of all approaches decrease. This is consistent with our commonsense that it is more challenging to generate a longer phrase. 

From the evaluation results in Table \ref{table2}, similar observations could be found. First, the proposed CVAE-KD achieves the best results on all criteria except for ROUGE-2 metric. The scores of ROUGE-1 and ROUGE-L of our model are respectively 9.5$\%$, 49.5$\%$ higher than that of the CVAE model. It is noticed from both tables that the original CVAE outperforms Seq2seq and Seq2seq+Attn. Apparently, the CVAE well fits for this problem as the CVAE-type models are generative based approaches which can decode a high-dimensional output data using the learnt low-dimensional latent variable. 
Furthermore, with the knowledge distillation and the employed pre-trained model, the proposed CVAE-KD could further enhance the model performance, and this verifies the effectiveness of the proposed approach. 



\subsection{How the size of financial reports affect model performance (RQ2)}

This experiment is to evaluate what is the model performance of the proposed approach when generating financial reports of different length. To recall that, we empirically set the length of the generated reports to 100, 150 and 200, respectively. We expect that the model performance of CVAE-KD will gradually decrease when generating longer reports. The corresponding experimental results are reported in Table \ref{tab:lengthBleu} and Table \ref{tab:lengthRouge}, respectively. 

For simplicity reason, we denote report(200) as the generated report with its length as 200 words, and similarly we have report(100) and report(150). For the BLEU evaluation criterion, as shown in Table \ref{tab:lengthBleu}, it is noticed that the BLEU-1, BLEU-2, BLEU-3 and BLEU-4 scores of report(100) are 2.5$\%$, 4.8$\%$, 9.1$\%$, 13.2$\%$ higher than that of report(150) which is the second best model. And the report(200) is the worst model w.r.t. all BLEU criteria. In addition, we also noticed that the BLEU-1, BLEU-2, BLEU-3 and BLEU-4 scores of all reports gradually decrease which is consistent with our expectation that the model might not be accurate enough to generate a longer phrase. 

Similarly for the ROUGE results, as shown in Table \ref{tab:lengthRouge}, we found that the ROGUE-1, ROUGE-2 and ROUGE-L scores of report(100) are 4.4$\%$, 12.9$\%$ and 9.9$\%$ higher than that of report(200) which is the worst model, and the ROUGE score also gradually decreases from ROUGE-1 to ROUGE-L. From these observations, we can conclude it is quite a difficult task to generate long text from a short text. The shorter the report length, the better the effect of the generating text. These objective evaluation results partially verify the effectiveness of the proposed knowledge distillation based approach. The following experiment will evaluate the subjective performance of the proposed approach. 

\begin{table}[t]
\centering
\begin{tabular}{m{1.1cm} p{1.25cm} p{1.25cm} p{1.25cm} p{1.3cm} }
\toprule[2pt] 
\centering
{Report Length} & \centering{BLEU-1} & \centering{BLEU-2} & \centering{BLEU-3} & \,{BLEU-4}
\\
\hline
\end{tabular}
\begin{tabular}{m{1.1cm} p{1.25cm} p{1.25cm} p{1.25cm} p{1.3cm} }
\centering{100} & \centering{51.06}  &\centering {24.10} &\centering {14.53} & \quad{8.66}\\ 
\centering{150} & \centering{49.83}  & \centering{23.00} & \centering {13.32} & \quad{7.65}\\ 
\centering{200} & \centering{46.67}  & \centering{20.32} & \centering{12.81}& \quad{8.00}\\
\bottomrule[2pt]
\end{tabular}
\caption{The BLEU results of the proposed CAVE-KD on generating reports of different length.}
\label{tab:lengthBleu}
\end{table}

\begin{table}[t]
\centering
\begin{tabular}{m{1.1cm} p{1.8cm} p{1.8cm} p{1.8cm} }
\toprule[2pt] 
\centering
{Report Length} & \centering{ROUGE-1} & \centering{ROUGE-2} & \,{ROUGE-L} \\
\hline
\end{tabular}
\begin{tabular}{p{1.1cm} p{1.8cm} p{1.8cm} p{1.8cm} }
\centering{100} & \centering{19.08}  &\centering {2.98} &\,\,\quad{7.64}\\ 
\centering{150} & \centering{18.59}  & \centering{2.91} & \,\,\quad{7.55}\\ 
\centering{200} & \centering{18.27}  & \centering{2.64} & \,\,\quad{6.95} \\
\bottomrule[2pt]
\end{tabular}
\caption{The ROUGE results of the proposed CAVE-KD on generating reports of different length.}
\label{tab:lengthRouge}
\end{table}

\subsection{A case study of generating financial report (RQ3)}
To evaluate the quality of the generated reports, we choose a pair of news-report data, and report the corresponding results by CVAE and CVAE-KD in Table \ref{tab:report}. Note that due to page limit, the length of generated reports in this table is set to 200. We translated the generated Chinese reports into English via google translator and highlight the correct words in blue color.  

From the generated results, we have following observations. First, the hit rate of CVAE-KD-report is higher than that of CVAE-report, which means that the CVAE-KD is more accurate than CVAE. Second, the generated reports by CVAE-KD is more coherent to the target report. For example, ``The growth rate of manufacturing investment has fallen'' generated by CVAE-KD-report is more accurate than ``the US inflationary pressure'' generated by CVAE. These two observations verify that the quality of the generated reports by CVAE-KD is satisfying. However, it is also noticed that there still exist some repeated words or sentences in the generated CVAE-KD-report, e.g. ``We believe that we believe that we will'' in the last sentence, which needs further research efforts. 

\begin{table*}[t]
\centering
\small
\begin{tabular}{p{4cm} p{12cm}}
\toprule[2pt] 
\centering{
Input News} & {From the perspective of the U.S.'s own situation, the U.S. economic prospects are improving but the economic rebound will be less than expected.Private consumption in the U.S. depends on employment conditions and wage growth. When the number of employees is at a high level, the unemployment rate has dropped.At the same time, weaker U.S. inflation will further slow down wage growth.} \\\hline
\centering{Target Report} & {Therefore, subject to wage growth and declining inflation, the recovery of private consumption in the United States will be less than expected.Secondly, in terms of private investment, US housing inventories are at a low level, and housing prices continue to rise, which will stimulate the rapid growth of new housing construction and real estate investment; but another On the one hand, from the perspective of the inventory cycle, the U.S. manufacturing inventory replenishment is nearing completion. It is expected that the monthly inventory cycle will peak and fall, which will drag down manufacturing investment and therefore the U.S. fixed asset investment in the third quarter. Performance will be difficult to regain the strong first quarter.}
\\ \hline
\centering{CVAE-report} & {"We believe that,, of,   month   month   day forecast; ())) month forecast; ((,,, and economic \blue{growth} expectations and economic The \blue{growth} rate fell back to---The \blue{economic} growth rate fell back to the US dollar index, which was a month-on-month decline; the US dollar index was a month-on-month decline and the \blue{US} dollar index was a quarter-on-quarter decline, and the year-on-year growth rate fell, but it will continue in the future.  \blue{the US inflationary pressure}; the market-to-market-to-market-to-China policy policy policy is also the government debt crisis risk appetite   under downward pressure   downward   is also the government To a certain extent.}\\\hline
\centering{CVAE-KD-report} & {In the economic data, the economic \blue{growth} month since the meeting is \blue{expected}. But the US \blue{economic} data in the \blue{United States} , the \blue{US government}, \blue{investment} and market expectations; while manufacturing investment growth rate down. \blue{The growth rate of manufacturing investment has fallen}. The \blue{growth} rate of \blue{manufacturing} is in line with "market expectations and" policies, etc. Under the policy: "Under the policy: The central bank’s meeting on the market’s inflation in the middle of last month: This will be the index to increase \blue{inflation}. Interest will be the main reason. The company will become a global enterprise and global enterprise field in the future. We believe that we believe that we will also have a global enterprise field in the future.} \\
\bottomrule[2pt]
\end{tabular}
\caption{The input news and the corresponding generated reports.}
\label{tab:report}
\end{table*}

\subsection{Results on ablation study (RQ4)}
Note that the proposed model contains several components. However, the most important one is the knowledge distillation (KD) component. To investigate whether the proposed knowledge distillation component works or not, we perform this experiment by removing the KD component and revising the model loss function. The comparison results are recorded in Table \ref{table6}. 

From this table, it is noticed that the model performance of CVAE with KD component is higher than that of the CVAE without KD component, especially for the BLEU scores. 
For BLEU-1, BLEU-2, BLEU-3 and BLEU-4 criteria, the score of the CVAE-KD is 9.9$\%$, 28.9$\%$, 44.4$\%$ and 64.6$\%$ higher than that of the CVAE without KD. Similarly, the ROUGE-1, ROUGE-2 and ROUGE-L score of the CVAE-KD is 18.8$\%$, 43.5$\%$ and 1.6$\%$ higher than that of the CVAE without KD, respectively. 
These results verifies that the proposed CVAE-KD could capture external knowledge to help the generation of financial reports to some extent. 
\begin{table*}[t]
\centering
\begin{tabular}{p{4cm} p{1.5cm} p{1.5cm} p{1.5cm} p{1.5cm} p{1.5cm} p{1.5cm} p{1.6cm}}
\toprule[2pt] 
\centering
{Methods} & \centering{BLEU-1} & \centering{BLEU-2} & \centering{BLEU-3} & \centering{BLEU-4} & \centering{ROUGE-1} & \centering{ROUGE-1} & {ROUGE-L}
\\
\hline
\end{tabular}
\centering
\begin{tabular}{p{4cm} p{1.5cm} p{1.5cm} p{1.5cm} p{1.5cm} p{1.5cm} p{1.5cm} p{1.6cm}}
\centering{CVAE-KD (without KD)} & \centering{42.46} & \centering{15.76} & \centering{8.87} & \centering{4.86} & \centering{15.38} & \centering{1.84} & \;\;\;\;{6.84}\\
\centering{CVAE-KD} & \centering\textbf{46.67} & \centering\textbf{20.32} &\centering\textbf{12.81} & \centering\textbf{8.00} & \centering\textbf{18.27} & \centering\textbf{2.64} & \;\;\;\;\textbf{6.95}\\
\bottomrule[2pt]
\end{tabular}
\caption{Results of the ablation study.}
\label{table6}
\end{table*}

\section{Related Work}
Existing text-to-text generation approaches could be classified into three categories, i.e., sequence-to-sequence based approaches, variational autoencoder based approaches and generative adversarial network based approaches, and we review these approaches in the following subsections. 

\subsubsection{Sequence-to-sequence based approaches}  In the literature, sequence-to-sequence based approaches have achieved superior model performance in various text to text generation tasks. Cho et al.\cite{cho2014learning} proposed a neural network architecture with RNNs as a sequence of encoding and decoding component. The encoder embeds the input sequence of variable-length into a fixed-length feature vector, then the decoder maps the vector back to the target sequence of variable-length. To generate long text, a Seq2Seq+Attn model \cite{bahdanau2014neural} is proposed which  
allows to search relevant word from the source sentences. Feng et al. \cite{feng2018topic} propose a multi-topic-aware long short-term memory (MTA-LSTM) network to generate a paragraph-level Chinese essay. The CopyNet model \cite{gu2016incorporating} incorporates the copying mechanism into the learning process of Seq2Seq model which achieves a better model performance. 
Similar attention based approaches could be seen in 
\cite{dong2017learning}. 
Recently, various transformer based approaches \cite{xu-etal-2020-self,koncel2019text,keskar2019ctrl} or pre-trained models \cite{du2020adversarial,Song2019MASSMS,dong2019unified,yang2019xlnet} are proposed and have achieved the SOTA performance in related tasks. 
And the financial report generation issue is first raised in \cite{hu2019automatically}. In their work, a two-stage hybrid deep learning model is proposed to generate macro research reports from a piece of breaking news. 

\subsubsection{Variational autoencoder based approaches} The variational autoencoder (VAE) models are also widely seen in text generation task. For instance, \cite{2015Generating} proposes a RNN-based VAE model which learns the feature representations of latent variables at the sentence level. The proposed model can explicitly represent the holistic properties of sentences such as style, topic, and high-level syntactic features.
An inference network \cite{miao2016neural} is proposed to apply on the discrete input to estimate the variational distribution. 
The authors \cite{miao2016language} further model the input text as a discrete latent variable under the variational auto-encoding framework. 
Then, a neural network-based generative architecture with latent stochastic variables \cite{serban2016hierarchical} is proposed to generate diverse text. To generate a long sequence of text, a hybrid architecture \cite{semeniuta2017a} is proposed to interweave the feed-forward convolutional and deconvolutional components. 
The conditional VAE model \cite{2019T} is considered as the state-of-the-art approach in this task. The CVAE employs a shared attention layer for both encoder and decoder, and this 
this is able to learn better feature representations of coherent sentences. 
Later, a multi-pass hierarchical CVAE \cite{yu2020draft} is proposed for automatic storytelling. Note that the CVAE-type models are generally resolved through the ELBO optimization \cite{mccarthy2020addressing}. 

\subsubsection{Generative adversarial network based approaches} 
The original generative adversarial network (GAN) \cite{goodfellow2014generative} is already widely adapted to various research problems. As the original GAN cannot model discrete variables, \cite{kusner2016gans} proposes to employ Gumbel-softmax distribution for this issue. To generate text, a LSTM module \cite{zhang2016generating} or GRU module \cite{zhu2018texygen} is commonly adopted as the generator component. By modeling the model generator as a stochastic process \cite{yu2017seqgan}, SeqGAN is proposed and is updated using a gradient policy rule. RankGAN\cite{lin2017adversarial} is then proposed to generate high-quality textual descriptions by revising the discriminator as a classifier. LeakGAN \cite{guo2017long} is further proposed to generate long text within 40 words. To reduce labeling cost, \cite{croce2020gan} proposes the GAN-BERT which extends the BERT-like architecture for modeling unlabeled data in a generative adversarial setting.


\section{Conclusion}
In this paper, we propose the conditional variational autoencoders based approach with knowledge distillation (CVAE-KD) to automatically generate long financial reports from a piece of short news. 
Particularly, a higher level latent variable is learnt from the background knowledge base respectively extracted for each input data. We then force the latent variable of CVAE to approximate  the higher level latent variable. At last, the knowledge distillation component is designed which takes the output of the pre-trained model as a teacher to better supervise the generation of financial reports. Extensive experiments are preformed on a public dataset to evaluate the model performance. The experimental results demonstrate that the proposed approach could achieve the state-of-the-art model performance against the compared approaches. 

\clearpage


\end{document}